\documentclass{article}

\usepackage[dblblindworkshop, final]{neurips_2026}

\workshoptitle{The 6th Workshop on Mathematical Reasoning and AI}

\usepackage[utf8]{inputenc}
\usepackage[T1]{fontenc}
\usepackage{hyperref}
\usepackage{url}
\usepackage{booktabs}
\usepackage{amsfonts}
\usepackage{nicefrac}
\usepackage{microtype}
\usepackage{xcolor}
\usepackage{xspace}
\usepackage{enumitem}
\usepackage{amsmath}

\newcommand{\AnnalsChallenge}{\textsc{AnnalsChallenge}\xspace}
\newcommand{\leaneval}{\textsc{LeanEval}\xspace}
\newcommand{\mathlib}{\texttt{Mathlib}\xspace}

\title{A dataset of formalised theorem statements from the Annals of Mathematics}

\author{
  Thomas Browning\thanks{Joint first author.} \\
  Department of Mathematics\\
  Imperial College London\\
  \texttt{t.l.browning@imperial.ac.uk} \\
  \And
  Katerina Hristova\footnotemark[1] \\
  Department of Mathematics\\
  Imperial College London\\
  \texttt{k.hristova@imperial.ac.uk} \\
  \And
  David Ledvinka\footnotemark[1] \\
  Department of Mathematics\\
  Imperial College London\\
  \texttt{d.ledvinka@imperial.ac.uk} \\
  \And
  Justus Springer\footnotemark[1] \\
  Department of Mathematics\\
  Imperial College London\\
  \texttt{j.springer@imperial.ac.uk} \\
  \And
  Hang Lu Su \\
  Department of Mathematics\\
  Imperial College London\\
  \texttt{hanglu.su@pm.me} \\
  \And
  Jack McKoen \\
  Department of Philosophy\\
  Carnegie Mellon University\\
  \texttt{mckoen@cmu.edu} \\
  \And
  Kevin Buzzard\thanks{Grant PI.} \\
  Department of Mathematics\\
  Imperial College London\\
  \texttt{k.buzzard@imperial.ac.uk} \\
}

\begin{document}
\bibliographystyle{plainnat}
\maketitle
\begin{abstract}
    Recent progress in autoformalisation of mathematics using AI has been impressive.
    Measuring this progress requires datasets of formal statements that are both hard and trustworthy.
    We present \AnnalsChallenge: 50 formalised theorem statements written by hand in Lean 4 using \mathlib.
    Each is the main result of a paper published in the \emph{Annals of Mathematics} since 2020.
    Every formalisation was manually reviewed and then audited by AI.
    We release them without formalised proofs.
\end{abstract}

\section{Introduction}

\AnnalsChallenge is a set of formalised theorem statements written in Lean 4
\citep{lean4}, using \mathlib \citep{mathlib2020}, its mathematics library.
These theorem statements are the main results of 50 papers published in volumes 191--203 (covering the period from 2020 to mid-2026) of the {\it Annals of Mathematics}, one of the top pure mathematics journals.
They constitute a quarter of the 207 papers published in this period.

The purpose of \AnnalsChallenge is three-fold: to analyse the state of \mathlib and its readiness for the formalisation of contemporary pure mathematics research, to enrich \mathlib with missing definitions and theorems, and to gauge the progress of AI autoformalisation applied to existing proofs.

For these reasons, \AnnalsChallenge differs from other existing datasets and is the first formalisation challenge of its kind for
contemporary pure mathematics research.
Competition-based datasets include {\tt miniF2F} (Olympiad-style problems from
the AMC, AIME, and IMO) \citep{minif2f}, {\tt FIMO} (IMO shortlisted problems)
\citep{fimo}, and {\tt PutnamBench} (from the William Lowell Putnam Competition)
\citep{putnambench}. Other datasets include {\tt FormalMATH}, spanning
high-school to undergraduate-level mathematics \citep{formalmath}, and {\tt
ProofNet}, targeted at undergraduate level mathematics \citep{proofnet}.
Domain-specific benchmarks like {\tt CombiBench} (combinatorics)
\citep{liu2025combibenchbenchmarkingllmcapability} and {\tt FATE} (algebra)
\citep{fate} focus on a single area of mathematics rather than reaching across a
wide spectrum of modern research mathematics. 

Research-level datasets spanning multiple areas of mathematics differ in their sources and construction.
{\tt ArXivLean} \citep{arxivlean,matharena} automatically extracts and
formalises statements from recent paper abstracts, followed by manual
review. Its formalisation pipeline is instructed to exclude candidates
requiring definitions not already present in \mathlib.
{\tt FormalConjectures} \citep{FormalConjectures} focuses on open problems without known informal
proofs, although  it also
includes solved research problems for proof autoformalisation. {\tt SorryDB} \citep{sorrydb} draws tasks from existing Lean formalisation
projects, asking models to supply missing proofs or fill gaps in partially
completed proofs.

\AnnalsChallenge was released on GitHub \citep{annalschallengegithub}, as well as on \leaneval, a public benchmark and leaderboard for autoformalisation of mathematics in Lean. On \leaneval, submitted proofs are checked for correctness without additional axioms in a sandboxed environment \citep{leaneval}. 

\section{Methodology}

\AnnalsChallenge has the following key design principles:

\begin{description}[leftmargin=0pt, font=\bfseries, topsep=2pt, itemsep=2pt, parsep=0pt]
    \item[Trust.] We use \mathlib whenever possible, where definitions have been reviewed by the community. When we introduce local definitions, they are carefully reviewed by humans and AI.
    \item[Faithfulness.] The formalised statement should be as close as possible to the informal source.
    \item[Correctness.] When the informal statement has gaps or is incorrect,
    we correct the statement to match the author's presumed intent.
    When the intended statement was unclear, we asked the authors of the paper.
\end{description}

\subsection{Statement selection process}

\label{sec:selection}

The first step of our selection process was a triage of papers according to
the difficulty of formalising their main theorems (statements, not proofs) in volumes 191 to 203. Excluding retractions,
errata and corrigenda, there are 207 papers published in this period. We
read each paper and identified the main result(s). We then assigned a difficulty
rating from 1 to 5, reflecting how hard we believe it would be to formalise the
statement given the current state of \mathlib. A rating of 1 means the statement
can be formalised directly with all relevant definitions already present in
\mathlib. A rating of 5 means the statement relies on definitions which would be
a major research project to formalise. A low rating does not imply that the theorem
is easy to prove.
For example, our dataset contains several theorems in finite group theory whose statements are elementary, but whose proof depends on the classification of finite simple groups.
The proof of the classification spans over ten thousand pages and has not been formalised to date.
Table~\ref{tab:difficulty} shows the distribution
of papers along our difficulty scale.

\begin{table}[t]
\centering
\caption{Distribution of papers by estimated difficulty of formalising
their main theorem statements.}
\label{tab:difficulty}
\begingroup
\begin{tabular}{c r r}
\toprule
Difficulty & Sample & Formalised \\
\midrule
1 & 40 & 36 \\
2 & 37 & 11 \\
3 & 56 &  3 \\
4 & 44 &  0 \\
5 & 30 &  0 \\
\midrule
Total & 207 & 50 \\
\bottomrule
\end{tabular}
\endgroup
\end{table}

We started with formalising the easiest statements while simultaneously aiming for a wide subject area coverage.

\subsection{Dataset statistics}
\label{sec:stats}

We took the subject classification of each paper from the {\it Annals of
Mathematics} website. Where it lists none, we fell back to MathSciNet. We count
both primary and secondary subjects, so one paper can count towards several
different areas. The rows in Table~\ref{tab:subjects} therefore sum to more than the
totals, which count each paper once.

\begin{table}[t]
\centering
\caption{Subject-area coverage of the 207 papers considered and the
50 papers whose main theorem statements were formalised. Papers may
belong to more than one subject area.}
\label{tab:subjects}
\begin{tabular}{lrrr}
\toprule
Subject area (MSC) & Papers & Formalised & \% \\
\midrule
Algebraic geometry & 54 & 8 & 14.8\% \\
Number theory & 50 & 20 & 40.0\% \\
Partial differential equations & 29 & 2 & 6.9\% \\
Differential geometry & 27 & 3 & 11.1\% \\
Dynamical systems and ergodic theory & 21 & 4 & 19.0\% \\
Group theory and generalizations & 20 & 8 & 40.0\% \\
Topological groups, Lie groups & 16 & 0 & 0.0\% \\
Several complex variables and analytic spaces & 15 & 2 & 13.3\% \\
Probability theory and stochastic processes & 14 & 4 & 28.6\% \\
Combinatorics & 13 & 8 & 61.5\% \\
Convex and discrete geometry & 10 & 6 & 60.0\% \\
Manifolds and cell complexes & 10 & 2 & 20.0\% \\
Other & 17 & 5 & 29.4\% \\
\midrule
Total & 207 & 50 & 24.2\% \\
\bottomrule
\end{tabular}
\end{table}

\subsection{Formalisation philosophy}

We formalised the statements manually over the course of ten months. The formalisations are built upon definitions and theorems in \mathlib, our only dependency. Wherever new definitions were needed, we either added them to \mathlib in the course of the project or, in the rare cases where a definition was paper-specific and of no broader interest, we developed it locally. Every formalised statement is kept as close to its natural language form as possible. We also preserved the generality of the original papers and followed their naming conventions and notation.

\subsection{Manual review}
After each statement was formalised, we performed an in-depth manual review checking for various types of issues: incorrect edge cases, inconsistencies with the paper, missing hypotheses, or errors in our interpretation of the paper. Our team included mathematicians with expertise across a range of fields and prior experience with Lean formalisation.
On a few occasions, we found issues with the statement in the paper, typically missing implicit hypotheses.
In such cases, we consulted the authors to confirm the intended interpretation and that the further hypotheses neither altered the meaning of their intended result nor its correctness.

\subsection{AI audit}
After the 50 statements were manually formalised and reviewed, we ran multiple AI audits on each file. We first used Claude Fable 5 (Max Effort), then GPT 5.6 Sol Ultra, and then Claude Fable 5  (Max Effort) again. 
The prompt, provided in Appendix A, was written to search for misformalisations. 
For 32 of the papers, the AI audit found no issues of mathematical significance.
For 7 of the papers, the AI audit revealed missing (mostly trivial) assumptions in the paper whose absence made the theorem false (see Appendix B for more details).
For the remaining 11 papers, the AI audit revealed mistakes in our formalisation.
We reviewed the AI findings manually and fixed all identified
issues before public release. Since release, no further misformalisations have been found. 

Misformalisation is a known difficulty in constructing datasets of formalised statements.
For example, as of 2 October 2026, the {\tt FormalConjectures} repository contains 3,604 formalised statements, and has had 511 misformalisation fixes merged.

\section{Lessons from the formalisation process}

To our surprise, we learned that mathematical practice in some areas of mathematics makes the work less amenable to direct formalisation. Authors sometimes write assuming conventions that may be clear to the experts in the field but are not explicitly stated. For example, implicitly assuming regularity criterion of certain objects such as measurability of functions or sets, or using  definitions such as the Fourier transform or solution to a PDE which can have different meaning depending on the context - the
definition of the Fourier transform could be an explicit function or distribution valued, and the definition
of the solution of a PDE could be a classical solution or various notions of weak solutions. Which definition is intended is not always clear from the definitions given explicitly in the paper and the main theorem. 

Going forward, a natural next goal would be to formalise the statements of the remaining 157 papers.
This is not as easy as it sounds; \mathlib is still missing many fundamental definitions and theories.
For example theorems about symplectic geometry, Riemannian geometry, K\"{a}hler geometry, Lie theory, intersection theory, infinity categories, moduli stacks, automorphic representations, and Shimura varieties are often essentially impossible to even state right now.
Getting the relevant mathematics into \mathlib will be a long process.

\section{Autoformalisation results}

\AnnalsChallenge has been live on \leaneval since 17 August 2026. 
Table~\ref{tab:leaderboard} shows the model standings as of 2 October 2026.
Accepted formal proofs had been submitted for 40 of the 50 challenges.
These results come from independent community submissions rather than a controlled evaluation: compute budgets, numbers of attempts, human involvement, and access to the source papers were not standardised across submitters.

\begin{table}[t]
\centering
\caption{Accepted community proof submissions on \leaneval as of
2 October 2026.}
\label{tab:leaderboard}
\begin{tabular}{lrr}
\toprule
Model/Harness & Proofs formalised / 50 & \% \\
\midrule
Axiom Prover (Axiom Math) & 40/50 & 80.0\% \\
Humanifa + GPT 5.6 sol & 34/50 & 68.0\% \\
NEAR AI w/ DeepSeek V4 & 33/50 & 66.0\% \\
Tau (caj.al) & 23/50 & 46.0\% \\
Midas & 20/50 & 40.0\% \\
\midrule
Total&40/50&80.0\%\\
\bottomrule
\end{tabular}
\end{table}

Autoformalisation was successful on 28 of the 36 difficulty level 1 papers, on
all 11 of the difficulty level 2 papers, and on one of the three difficulty
level 3 papers (see Section~\ref{sec:selection} for our difficulty scale). Note that the
difficulty range of the proofs is quite big, and may be uncorrelated with the
difficulty of formalising the statement. Some papers use rather elementary proof
techniques, while some others rely on large amounts of deep modern mathematics. 

\section{Conclusion and future work}

\subsection{Conclusion}

In this work, we presented \AnnalsChallenge, a novel dataset of theorem
statements in Lean~4. Unlike previous datasets, we cover a broad range of
subject areas from modern research level mathematics.

\mathlib is beginning to be ready for mathematics at this level: we were able to
state fifty recent main results on top of it, even if the difficulty
distribution in Section~\ref{sec:selection} shows how much is still out of
reach. Within weeks of release, more than half of these statements had a
formalised proof submitted to \leaneval, which we read as a sign of strong
interest in the dataset, and as an indication that autoformalisation will play
an important role in how mathematics is reviewed in the future.
\subsection{Future work}
We see three directions.
The first is to state more theorems, especially in the areas underrepresented in Section~\ref{sec:stats}.
Several of these areas are barely covered by \mathlib, so this would require building out the library first.
The second is to scale up with AI, for instance, by relying on definitions from \texttt{Tau Ceti} \citep{TauCeti}, an AI-managed library.
Every definition would still need
careful review, or the dataset would lose its trustworthiness.
The third is to cover
other journals, whose subject balance differs from that of the {\it Annals of Mathematics}.

\section{Acknowledgements}

This project is funded by Renaissance Philanthropy's 2025 AI for Math Fund.
We thank Renaissance Philanthropy and XTX Markets, the founding donor of the AI for Math Fund, for their support of this project.

We would like to thank Bhavik Mehta, Heather Macbeth and Mathias Stout for their help with formalising and reviewing the statements. We thank Kim Morrison for including our dataset into \leaneval.
Finally, we thank the authors who answered questions about their work:
Alex Cohen, Tuan Ngo Dac, Andrei Jaikin, Victor Lie, Andrew Lobb, Ezequiel Maderna, James Maynard, Bhargav Narayanan, Huy Pham, Anthony Poëls, Alex Smith, and Andrea Venturelli.

\bibliography{refs}

\appendix

\section{AI audit prompt}
The prompt for the AI audit was as follows:
\begin{quote}
    We are working on a project to formalise (in Lean) the statement of the main theorems from various Annals of Mathematics papers. We are focussing on just one paper here. The pdf of the paper is in the pdf directory, and the corresponding formal statement is proved in the file <FILE NAME>. The statements of the main sorried result or results in Lean should be equivalent to the corresponding statements in the paper, and as faithful to the statement in the paper as is reasonably possible. Please review the formalisation and docstring for any errors. In particular you should look for subtle errors in the theorem statement (including any definitions that they rely on) that make them wrong or unfaithful to the statement in the paper. You may want to run tests by checking special cases of definitions, or by proving or disproving special cases of the theorems. Please do not attempt to fill in the sorry following the paper -- that would be a major project! But if you can find a simpler proof or disproof of one of the sorries, thus indicating a misformalization, then feel free -- this would be a wonderful outcome. We are just auditing the formalization, not filling in the sorries from scratch. Please do not ask any questions when working, but instead make commits to the <BRANCH NAME> branch directly. Humans will review the suggestions afterwards. Do not push anything!
\end{quote}

\section{Missing or implicit hypotheses in papers}
We record the (essentially all trivial) slips in the published versions of the theorems
we formalised.
\begin{enumerate}
    \item\textit{Lorentzian polynomials} \citep{MR4172622}, Theorem 2.25: The paper allows $n$ and $d$ to be nonnegative, but the theorem is false for $n=0$ and $d=2$, where
    \[\mathring{L}_0^2=\{f\in P_0^2\mid\mathcal{H}_f \text{ is nonsingular and has exactly one positive eigenvalue} \}=\emptyset,\]
    since the $0\times0$ Hessian matrix of $f$ cannot have a positive eigenvalue, but $L_0^2=S_0^2$ contains the zero polynomial, since the zero polynomial is homogeneous of degree $2$.
    \item\textit{Thresholds versus fractional expectation-thresholds}
        \citep{MR4298747}, Theorem 1.1: The assumption that $\ell(\mathcal{F}) \geq 2$ is not stated explicitly in the paper, but it is necessary: if $\ell(\mathcal{F}) = 1$, then $\log(\ell(\mathcal{F})) = 0$, whereas $p_c(\mathcal{F}) \in (0,1)$, so the inequality cannot hold.
    \item\textit{On a conjecture of Talagrand on selector processes and a
        consequence on positive empirical processes} \citep{MR4740540}, Theorem 1.2: The assumption that
    \[0<\mathbb{E}\sup_{\lambda\in\Lambda}\sum_{i\in X_p}\lambda_i<\infty\]
    is not explicitly stated in the theorem but used in the last step of the proof where the inequality $\mathbb{E}\geq(L/L')\mathbb{E}$ is a contradiction for $L>L'$.
    \item\textit{On a conjecture of Talagrand on selector processes and a
        consequence on positive empirical processes} \citep{MR4740540}, Theorem 1.3: The requirement that the collection $\mathcal C$ be finite is not explicitly stated in the paper.
    However, $\mathcal C$ must at least be assumed to be countable in order for the sum to make sense,
    and the collections $\mathcal C$ produced by the proof are indeed finite.
    \item\textit{Algebraic integers with conjugates in a prescribed
        distribution} \citep{MR4768417}:
    The definition of $\lambda_{SSS}$ contains a typo that was caught during human review.
    The paper said ``least real number'',
    but this is clearly a typo, as this set does not have a lower bound.
    Defining it as the greatest real number is consistent with how this constant
    is defined in another paper by the same author
    \citep{S0025-5718-2024-04004-8}.
    \item\textit{Symplectic monodromy at radius zero and equimultiplicity of
        $\mu$-constant families} \citep{MR4768419}, Theorem 1.1: We need to require the constant coefficients to vanish. This is equivalent to saying that $f_t$ is a germ $(\mathbb{C}^n,0)\to(\mathbb{C},0)$, i.e. that it is a family of isolated hypersurface singularities through the origin, which is mentioned both in the abstract and in Corollary 1.3, but not explicitly in Theorem 1.1.
    \item\textit{On a conjecture of Marton} \citep{MR4880432}: The set $A$ must be nonempty, but this is left implicit in the paper.
    \item \textit{Galois groups of random integer polynomials and van der
        Waerden's Conjecture} \citep{MR4878222}, Theorem 1: We must impose $n\geq3$, as the result is false as stated for $n=2.$
    \item\textit{Fractal uncertainty in higher dimensions}
        \citep{MR4927737}, Theorem~1.1: The sets $X$ and $Y$ are intended to be measurable, but this is not explicitly stated.
    This was caught during human review and clarified with the authors.
    \item\textit{New large value estimates for Dirichlet polynomials}
        \citep{MR5037841}, Theorem~1.1: The inequality may be false at $T=1$, so we must impose $T>1$.
\end{enumerate}

\end{document}